\documentclass[letterpaper, 10 pt, conference]{ieeeconf}\usepackage{graphicx}  

\IEEEoverridecommandlockouts                              

\usepackage[T1]{fontenc}
\usepackage{caption} 
\usepackage{array}
\usepackage[table,svgnames]{xcolor} 
\usepackage{booktabs}
\usepackage[pagebackref,breaklinks,colorlinks,bookmarksopen,bookmarksnumbered,citecolor=red,urlcolor=red]{hyperref}
\usepackage[capitalize]{cleveref}
\crefname{section}{Sec.}{Secs.}
\Crefname{section}{Section}{Sections}
\Crefname{table}{Table}{Tables}
\crefname{table}{Tab.}{Tabs.}

\usepackage[colorinlistoftodos, disable]{todonotes}
\usepackage{comment}

\newcommand{\wenzhen}[1]{\todo[inline,color=red!40]{#1}}
\newcommand{\dakarai}[1]{\todo[inline,color=green!40]{#1}}

\newcommand{\maria}[1]{\todo[inline,color=purple!40]{#1}}

\title{\LARGE \bf
Social Gesture Recognition in spHRI: Leveraging Fabric-Based Tactile Sensing on Humanoid Robots
}

\author{Dakarai Crowder$^{1}$, Kojo Vandyck$^{1}$, Xiping Sun$^{1}$, James McCann$^{2}$, and Wenzhen Yuan$^{1}$
\thanks{$^{1}$Dakarai Crowder, Kojo Vandyck, Xiping Sun, and Wenzhen Yuan are with the University of Illinois at Urbana-Champaign, Champaign, IL, USA 
        {\tt\small \{dcrowd3,vandyck,xipings2,yuanwz\}@illinois.edu}}%
\thanks{$^{2}$James McCann is with Carnegie Mellon University, Pittsburgh, PA, USA
        {\tt\small jmccann@cs.cmu.edu}}%
}

\begin{document}

\maketitle
\thispagestyle{empty}
\pagestyle{empty}
\begin{abstract}
Humans are able to convey different messages using only touch. Equipping robots with the ability to understand social touch adds another modality in which humans and robots can communicate. In this paper, we present a social gesture recognition system using a fabric-based, large-scale tactile sensor placed onto the arms of a humanoid robot. We built a social gesture dataset using multiple participants and extracted temporal features for classification. By collecting tactile data on a humanoid robot, our system provides insights into human-robot social touch, and displays that the use of fabric based sensors could be a potential way of advancing the development of spHRI systems for more natural and effective communication.

\end{abstract}
\section{INTRODUCTION}

\wenzhen{Need to clarify the major contribution}
Humans interact with each other using many differing modalities and touch is one that occurs naturally. Social touch serves many purposes, such as fostering attachment between humans, allogrooming, and facilitating communication \cite{Suvilehto2023}. For example, a handshake might express gratitude \cite{Thompson2011}, while a gentle stroke conveys love or warmth \cite{Hertenstein2006}. As robots are increasingly deployed everywhere \cite{Lum2020}, including the deployment of humanoid robots in public spaces \cite{Nguyen2020}, enhancing a robot's sociability through touch \cite{Mazursky_22} could help foster natural interactions between humans and robots. This requires robots to be equipped with large-scale tactile skins and the ability to accurately recognize and respond to various social-physical human-robot interactions (spHRI).

\begin{figure}[t]
    \centering
    \includegraphics[width=\linewidth]{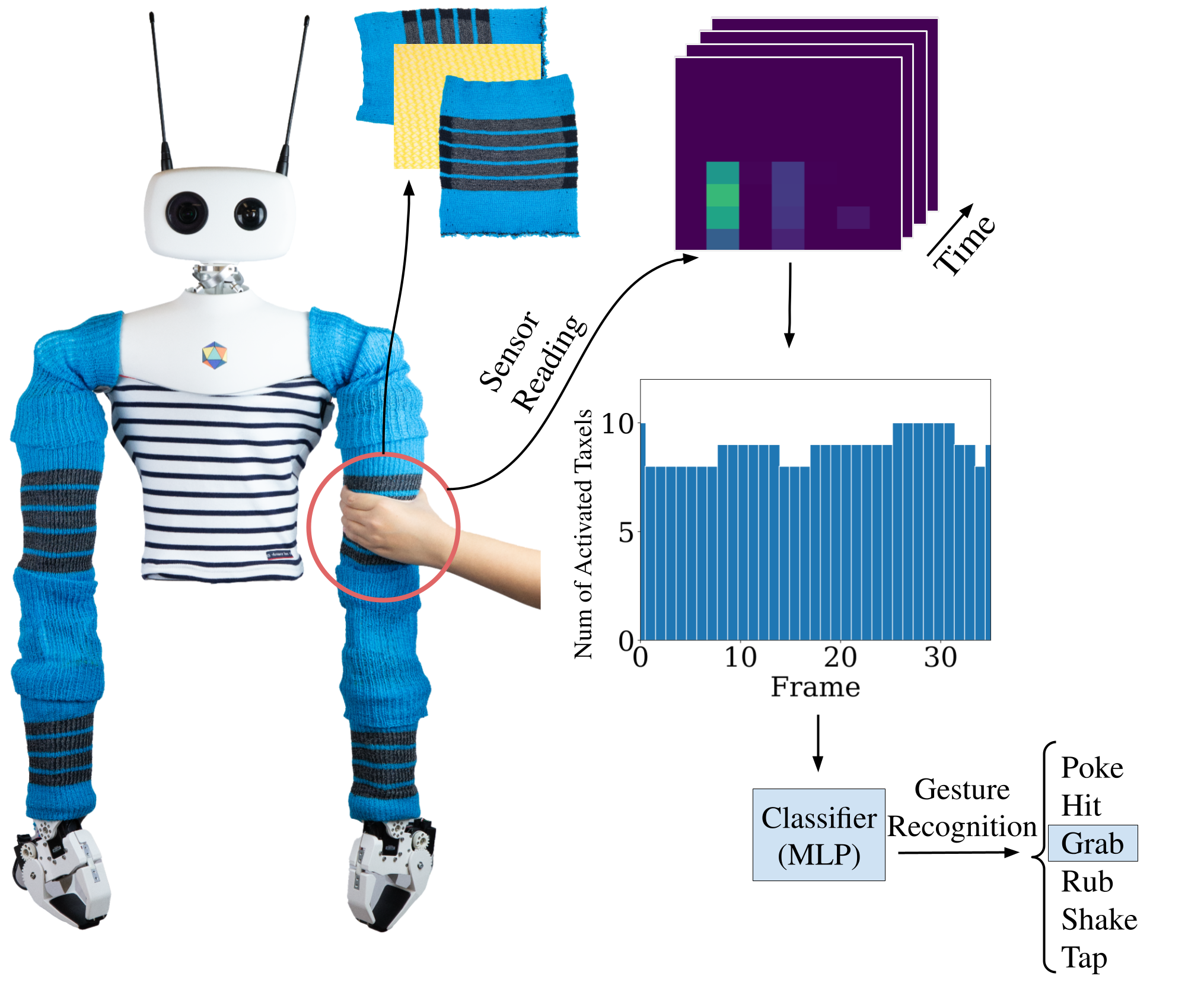}
    \caption{Our machine knitted large scale sensor has the ability to detect different types of gestures. Features are extracted from the sensor reading and fed into a model to classify the gesture being performed.}
    \label{fig:teaser}
\end{figure}

\wenzhen{In this paragraph, you need to better explain the difficulty of applying exiting tactile skins}

Despite the importance of having an effective system for humanoid robots to recognize social gestures, a good solution remains elusive. Many gesture classification studies rely on classifying data which was not collected on a humanoid robot \cite{Sarwar21,Albawi2018-on}. Those that do involve robots often employ small pressure sensors or miniature robots \cite{li2022ie, Montero2020}. Although there has been a lot of work done on the creation of robotic skins, the existing work tends to be low resolution, time-consuming to fabricate, or hard to customize for different robot morphology \cite{Pang21,yang24,park244}. With these limitations it is hard to use those skins on humanoids for spHRI.

\wenzhen{The following two paragraphs are not well organized. I suggest you to start with one or two sentences summarizing what you create in this paper. Then you can talk about the technological details, the advantage and influence of your technology, etc.}

In this paper, we introduce a system for social gesture recognition that incorporates a fabric-based, large-scale tactile sensor attached to the arms of a humanoid robot. To classify gestures, we built a social gesture dataset by collecting data from various participants and extracting temporal features for analysis. Our sensor was machine-knitted and designed to maintain accuracy even when the robot is in motion. The flexibility of the machine-knitted sensor enables easy customization to fit different shapes, while data collection on an actual humanoid robot offers insights into how humans interact with robots through touch. We believe that developing better systems for gesture recognition will allow for more natural and enjoyable social-physical human-robot interaction.

\wenzhen{Talk more about the contribution in gesture recognition}

\section{Related Works}
\subsection{Sociability of Robots}
Incorporating touch into human-robot interactions has the potential to enhance the perceived sociability of robots. Research has shown that anthropomorphizing social robots positively influences how people engage with them \cite{Roesler2021-gi}. By expanding a robot's social capabilities, it can be perceived as more approachable and friendly. This phenomena is often associated with the robot's ability to understand and respond to various human social interactions \cite{Breazeal2003-nl}. Giving robots human-like social skills contribute to a greater sense of familiarity and comfort, which can foster trust \cite{Saunderson2019-fh}. There is also a positive correlation between robot sociability and robot trust \cite{Kadylak2023}. Touch, as a fundamental mode of human communication, could further enhance this sense of sociability. By incorporating touch into interactions, robots may appear more relatable and human-like, which could facilitate deeper social engagement.

\subsection{Social Physical Human Robot Interaction}
Different modalities are often used to study and improve human-robot interaction (HRI). While audio \cite{Simpson2020, Baecker2020, El_Kamali2020} and vision \cite{Robinson2022-el, gao2022-ee, Li2020-xi, Fan2022-fy} are popular, we focus on social touch. Research shows that when people perceive robots as sociable, their trust in the robot increases \cite{Kadylak2023}. Social touch may be one way to enhance this perception. For example, studies on robot-initiated intra-hug gestures have shown that such actions foster an environment of understanding and comfort between humans and robots \cite{Block2019-xo, Block2021-gp, Block2023-bw}. Playful physical interactions, such as hand-clapping games, have also been studied for their impact on human-robot connection \cite{Fitter2020-di, Fitter2018-oa, Fitter2016-xm}. Additionally, physical touch has been shown to significantly enhance the quality and engagement in social-physical exercises with robots \cite{Fitter2020-di}.

Humanoid robots can assist in various settings, including healthcare  \cite{Papadopoulos2020}, education \cite{Ekstrm2022}, and companionship \cite{Papadopoulos2020}. So we implemented our system on Pollen Robotics Reachy 2023 (\cref{fig:teaser}). The robot's arm has seven degrees of freedom and they tend to mirror the arms of a human. The robot also has more of a cartoon looking face making the robot have a more friendly look. 

\subsection{Full body robotic skins}
One of the main objectives of a robotic skin is to cover as much of a surface as possible. There are many differing methods used to create robotic skins. One way to create the skin is to make a grid of tactile sensors \cite{Lin2021-ih,Pang21}. The matrix is formed using orthogonal strips of electrodes. Other methods tend to use multiple different cells and attach them together to make their sensor \cite{Bhattacharjee13, Schmitz11}. Robotic skins with the intention of being used with humans are created in a way where the sensor is comfortable to touch and or has the ability to detect different types of gestures. Mimicking the functionality of human skin through the use of microphone nodes \cite{Park2022-km,yang24} is one method used. Some works created a soft pressure robotic skin using 3d printed parts for safe interaction between humans and robots \cite{park244}. CushSense is another sensor which is a capacitive based sensor with an application of moving human limbs in a medical setting \cite{xu2024cushsense}. There are also many works done of fabric based sensors \cite{Inaba, Luo2021, Luo2023}. Our sensor is a continuation of the work done by Si \cite{Si2023}. RobotSweater is a machine knitted resistive-based sensor that has the ability to localize contact and obtain pressure information.  We are taking this working and increasing the scale of the sensor while also testing the applications for gesture recognition.
\maria{Any related work on resistive based sensors? Why are resistive based sensors a better option than existing sensors?}

\subsection{Social Gesture Recognition}
Enabling gesture recognition in robots offers a promising way for them to interpret messages communicated through touch. Humans naturally convey and interpret various emotions through tactile interactions \cite{Hertenstein2009}, and researchers have been investigating how these emotional expressions translate to human-robot touch communication \cite{Andreasson2018}. For instance, Yohanan and MacLean examined the types of gestures people use to express specific emotions \cite{Yohanan2012-ub}. They also developed a gesture touch dictionary inspired by human-animal and human-human touch, as well as affective touch patterns \cite{Yohanan2012-ub}. Understanding touch gestures presents an opportunity to create an additional communication modality between humans and robots, driving research into gesture recognition. One notable resource is the CoST dataset, which provides pressure data from various social gestures performed on a sensor \cite{Jung2014}. Some studies have focused on extracting features from such gesture datasets, including spatiotemporal characteristics, for classification tasks \cite{li2022yk, li2022ie}. Recent work has also explored incorporating shear forces into touch gesture recognition, resulting in improved classification accuracy \cite{choi2022}.

Although the CoST dataset offers a valuable foundation by capturing diverse pressure data from individuals performing different social gestures, it does not involve humanoid robots, which may evoke different gesture behaviors. Furthermore, the sensor used in the CoST dataset differs from the one employed in our research, preventing direct application of their dataset.

\section{Sensor Design and Fabrication}

Our sensor is specifically designed for social physical human-robot interaction and was mounted onto a humanoid robot, as this is its intended use. We opted for machine knitting the sensor due to its scalability and fabric's familiarity to humans. This work builds upon the foundation laid by Si \cite{Si2023}, maintaining the same working principles and overall design.

We knitted a three-layered, resistive-based tactile sensor. The upper and bottom layers have horizontal and vertical conductive stripes respectively (\cref{fig:teaser}), while the middle layer consists of an insulating mesh. The sensor’s resistance changes in response to fabric deformation \cite{Si2023}. When no pressure is applied, the upper and lower layers do not make contact. As pressure increases, the contact area between these layers increases, resulting in a decrease in resistance.
\maria{Information in the above paragraph would fit better in section II}

\begin{figure}
    \centering
    \includegraphics[width=8cm]{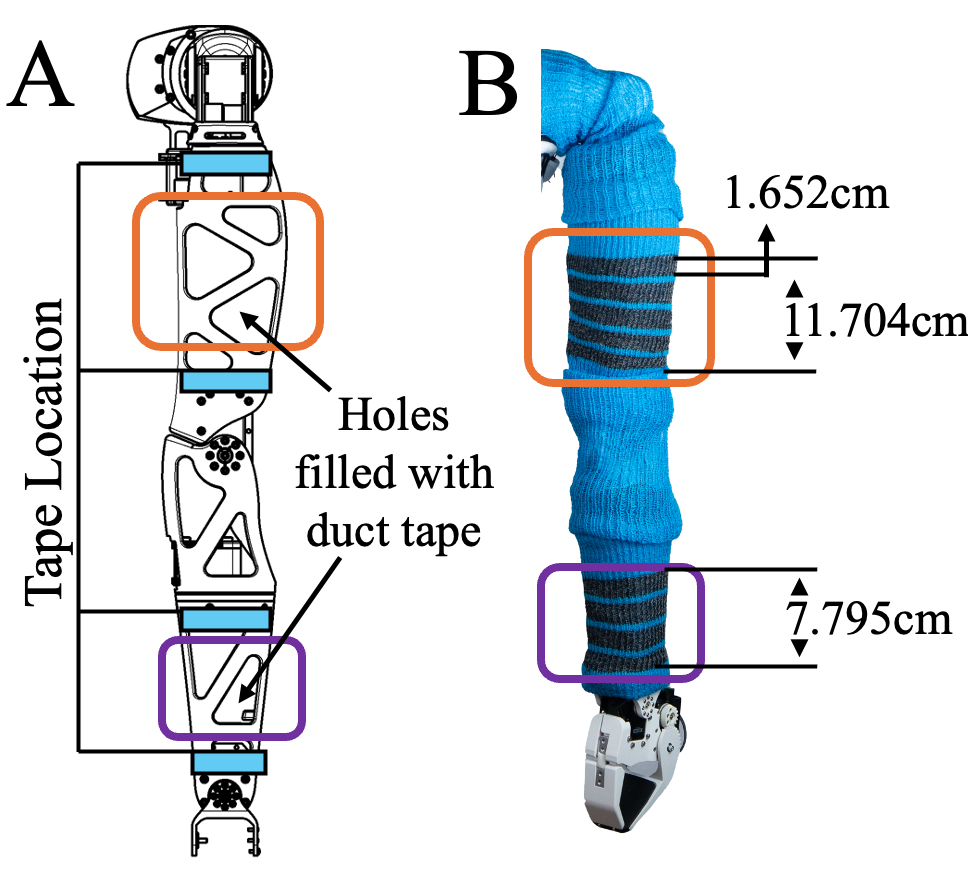}
    \caption{The model \cite{pollen_robotics_documentation} and actual pollen robotics arm. (A) The location that the double sided tape was placed and the holes which were filled with duct tap. (B) The sensor attached to the arm with corresponding sensing area measurements. The area where the horizontal stripes span on the upper and lower arm(the orange and purple box) are the sensing area.}
    \label{fig:sensor_fab}
\end{figure}

\begin{figure}
    \centering
    \includegraphics[width=\linewidth]{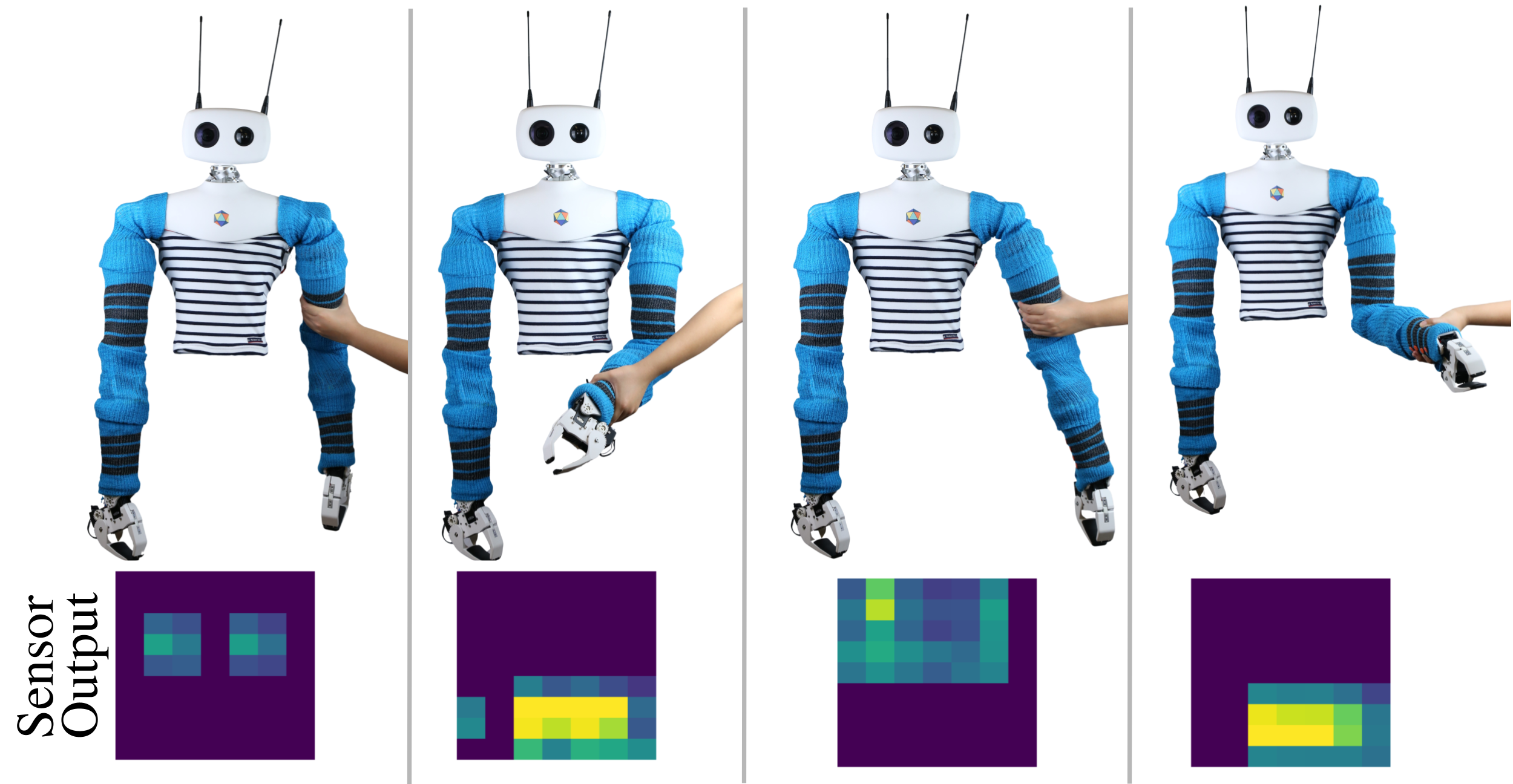}
    \caption{Reachy's arm in differing positions and the corresponding sensor reading. }
    \label{fig:reachy_in_dif}
\end{figure}

We attached the sensor to the robot in a way which prevents the fabric from moving while the arm is in motion. Our sensor is composed of two components. One is affixed to the upper arm of the robot, while the other is attached to the lower arm (\cref{fig:sensor_fab}). In the figure, the area where the black strips span is the sensing area. The blue portion which is covering the joints is there for aesthetic purposes only. This means that the elbow region of the arm is not covered by sensors. This area of coverage is to be done in future works. EZlifego double sided was used to fix the sensor on to the robot. Each sensing part of the sensor (the blue and purple boxes in \cref{fig:sensor_fab} (B)) was attached to only one link, to prevent the sensor from giving readings when the robot's arm was in motion \cref{fig:reachy_in_dif}. The robot has holes in the arm (\cref{fig:sensor_fab} (A)) which needed to be addressed. Using duct tape, the holes where covered, but this caused a decrease in sensor performance. 

The upper portion of the sensor has a grid of 5 columns by 7 rows and the lower sensor consists of a 4x7 grid. The sensor can distinguish between varying pressure levels and accurately localize contact with a spatial resolution of 1.652cm\textsuperscript{2}. The sample rate was around 50 Hz.
 
The sensor is composed of three machine knitted layers. Following the design from Si\cite{Si2023}, there are two conductive layers which sandwich an insulating mesh layer between them. The sensor was designed using Shima Seiki's KnitPaint and the final sensor was around 6mm thick.

The conductive layers of the sensor (the outermost and innermost layers) were knitted using a 1x1 rib pattern. A rib pattern allows for the fabric to stretch and conform to non uniform surfaces. Conductive strips were composed of acrylic yarn and Baekert BK 9036129, a Bekinox-polyester blend in 50/1 NM (the black strips in \cref{fig:teaser}. The Baekert yarn, which is the conductive yarn, was plated on the outside of the acrylic yarn. The yielded resistance was around 7-10$\Omega$/cm. The insulating portion, separating each of the conductive strips, was knitted with acrylic yarn (Tamm Petit 2/30). The stitch number, how large the stitch is, used on the insulating and conductive strips on these layers was 33. The insulating mesh between the conductive layers was knitted with a porous lace pattern. Acrylic yarn was used to form the mesh.
\maria{Just to be clear, the 'conductive portion' of the conductive layers are the black stripes? What is stitch number? Is that stitches per inch? What do you mean that the conductive portion was "pleated on the outside"? For the conductive layers, was the acrylic yarn for the light blue non-conductive part and the Baekert yarn for the black striped part? What is the "lace pattern" on the insulating mesh? Also, I assume you mean 'insulating' instead of 'insulting'.}

\section{Gesture Recognition}

\begin{figure*}[ht]
  \centering
   \includegraphics[width=\linewidth]{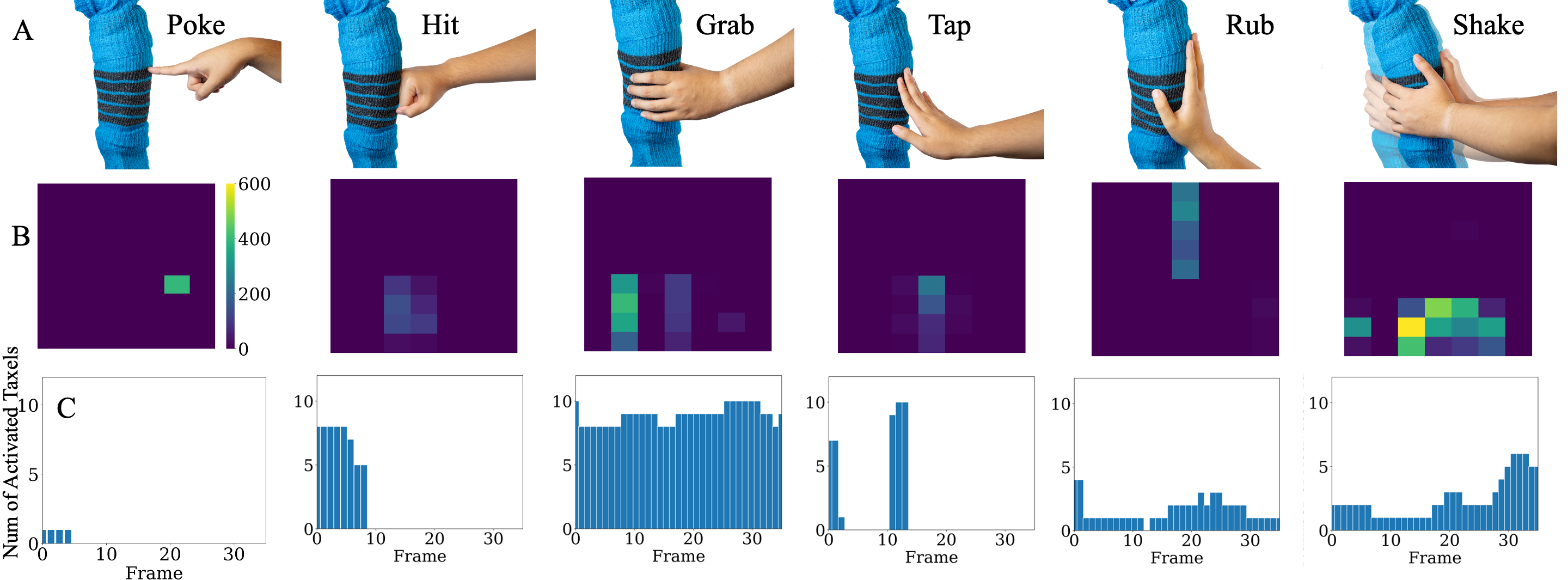}
   \caption{Display of the gestures and the signal reading. (A) A visual of the gesture being performed on the upper portion of the sensor. (B) A grid displaying what the raw signal looks like. The values shown are the mean taxel value. (C) the feature extracted from the sensor reading, the number of taxels activated per frame.}
   \label{fig:fullgestures}
\end{figure*}

In this section, we introduce our algorithm for recognizing social gestures based on the tactile data collected from the skin. We first heuristically extract a set of low-dimentional features from the spatial-temporal input of tactile signals and then use a Multi-Layer Perceptron (MLP) classifier to recognize the gestures. 

\subsection{Gesture Selection}
\begin{table}[b]
\caption{Gestures}
\label{tab:gesture}
\centering
\renewcommand{\arraystretch}{1.5}
\begin{tabular}{m{10mm} p{68mm}}
\toprule 
\textup{\textbf{Gesture}}         & \textbf{Description}\\
\midrule
Hit                      & Deliver a forcible blow to the arm with a closed fist\\
Poke                     & Jab or prod the arm quickly with one of the fingers\\
Grab                     & Grasp or seize the arm suddenly and roughly with whole curved hand\\
Rub                      & Move your hand repeatedly back and forth on the arm with firm pressure, lasting around 5 seconds\\
Shake                    & Move the arm up and down or side to side with rapid, forceful, jerky movements\\
Tap                      & Strike the Haptic Creature with a quick light blow or blows using one or more fingers multiple times\\
\bottomrule
\end{tabular}
\end{table}
Using the touch dictionary \cite{Yohanan2012-ub}, we choose six gestures for our algorithm to classify (\cref{tab:gesture} and \cref{fig:fullgestures}). We only selected a subset of the touch dictionary because some gestures do not adapt well to humanoid robots. For example, gestures like "finger idly" and "cradle" are only suitable for animal-like robots, so they were excluded. Additionally, some gestures were omitted because they generated similar signals from our sensor, making them difficult to distinguish. For instance, "pat" and "tap" produced nearly identical readings, and even humans may struggle to differentiate between them, so they were not included in the final selection.

\subsection{Data Prepossessing and Feature Extraction}
\wenzhen{The content is not clear. Please be more specific of the method you used, and explain the details.}
We preprocessed and extracted temporal features from our 3D spatial temporal data. We removed any frames recorded before the initial contact on the sensor. This was necessary because sensor output was recorded before interaction occurred. The feature extracted was the number of taxels activated per frame. An activated taxel is defined by having a digital reading which is larger than 10. For each frame, we summed the number of taxels that were activated. This resulted in an array of numbers representing the number of taxels activated for every frame. Each gesture varied in frame length so the data was either padded or clipped to be 150 frames. The number of taxels tended to be unique for each gesture. Other features, like pressure information or more spatial features did not yield a higher accuracy. 

\subsection{Model Architecture}
Our classifier used for gesture recognition was a four-layered MLP of size 256x128x64x6. We used a MLP because compared to other models tried it had the least amount of overfitting. Our training dataset only contains 900 gestures making models prone to overfitting. We initially used Random Forest because of our dataset size. Although it did have similar results as MLP, MLP still out performed.

\section{Experiments}
In this section, we introduce our experiments on characterizing the tactile skin and recognizing gestures. 
\wenzhen{The organization of each section still needs improvement}
\subsection{Sensor characterization}
\begin{figure}
    \centering
    \includegraphics[width=7.4cm]{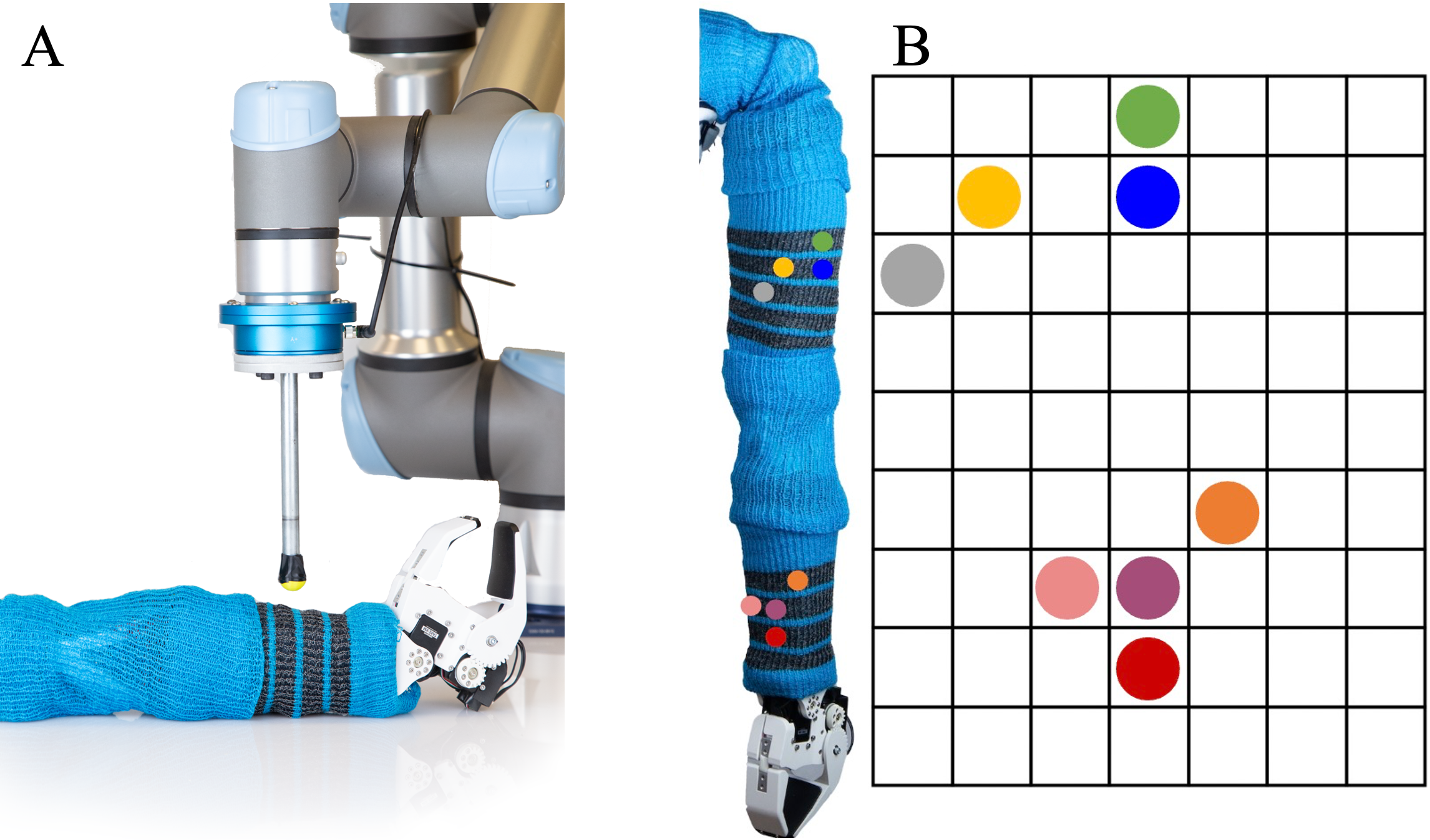}
    \caption{(A) Indenter (yellow semi sphere) with diameter of 20.25mm used (B) Taxels used for characterization experiments.\wenzhen{Add more space between two figures}}
    \label{fig:SensorPosition}
\end{figure}

For the sensor characterization tasks, we used a UR5e robot with a semi-spherical indenter attached to the end-effector (\cref{fig:SensorPosition} (A)). The indenter was mounted onto a Nordbo NRS-6050-D50 force-torque sensor, which provided ground truth force measurements. The robotic skin was fixed to Pollen Robotics' Reachy 20223 (\cref{fig:sensor_fab}).

\begin{figure}
    \centering
    \includegraphics[width=\linewidth]{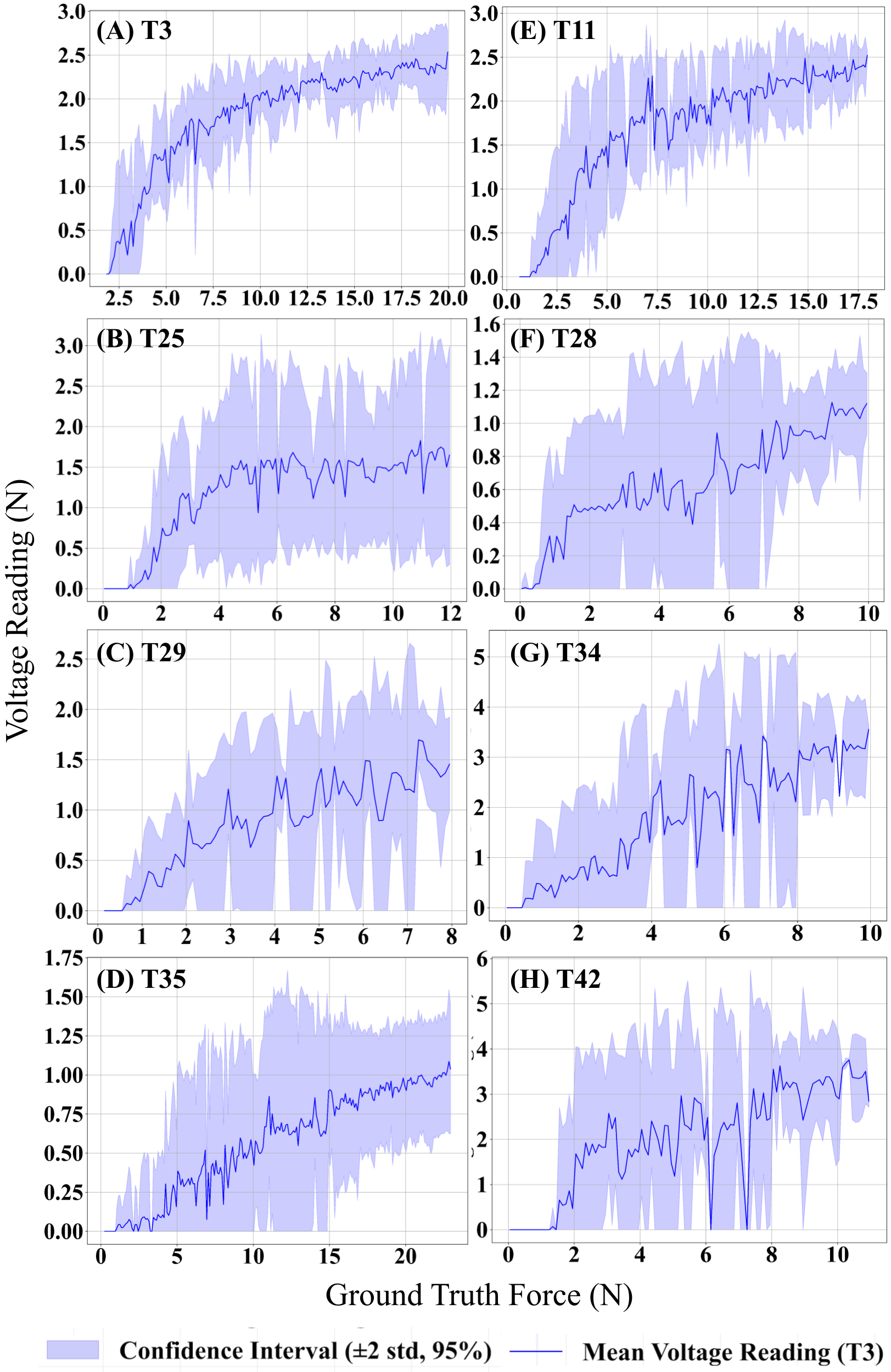}
    \caption{Voltage reading vs. Ground Truth for eight different taxels. (A), (C), (E) and (F) are taxels from the upper arm. (B), (D), (G), and (H) are taxels from the lower arm}
    \label{fig:characterization}
\end{figure}

We quantified the lowest and highest force measurable by the sensor for 8 different taxels (\cref{fig:characterization} (A)). To measure this range, the indenter was positioned 6cm above the tactile skin. The UR5e robot moved towards the sensor at a speed of 17mm/second taking around 35 seconds to make initial contact. Once contact is made the UR5e continues going down at the same speed for 4mm. Then the indenter was risen. The task was conducted ten times for each taxel (\cref{fig:SensorPosition} (B)). 

\dakarai{add the sensitivity range}

\wenzhen{Start this paragraph with some more positive conclusion. First describe the most important feature you learnt from the experiment}

The upper and lower portions of the sensor had slightly different sensing ranges. The upper sensor's range was 1.15 N (± 0.728) to 13.95 N (± 5.099), where the first value represents the lowest measurable force and the second represents the highest. The lower sensor's detection range was 1.975 N (± 0.492) to 13.95 N (± 5.244).

The readings from our sensor were not entirely consistent. As shown in \cref{fig:characterization}, some taxels exhibited a more linear relationship (\cref{fig:characterization} (G)) with the applied force than others (\cref{fig:characterization} (A)). We observed that the sensor performed more consistently when placed on a hard surface. However, when positioned on a more deformable surface, like the tape used to cover holes in the arm (\cref{fig:sensor_fab} (A)), the sensor's readings were less consistent. While the sensor demonstrated decent accuracy, it had low precision. Each taxel also behaved differently, even when placed on surfaces of similar hardness. This could be due to inconsistencies in the amount of resistive yarn or variations in the stretch across different parts of the sensor. Additionally, since the robot's arm is not uniform in radius, certain areas experienced more strain than others. Despite these challenges, we were able to classify most of the different gesture types.

\subsection{Data Collection for Gesture Recognition}
We collected data from multiple volunteers to create a social gesture dataset, consisting of six distinct gesture classifications (\cref{tab:gesture}), using our sensor. Afterward, we extracted various features from the data and trained several models to determine the most effective method for classifying the gestures.

\subsubsection{Setup}
Our setup consisted of the sensor mounted on Pollen Robotics' Reachy 2023 and a monitor. The upper section of the sensor had a 5x7 grid, while the lower section featured a 4x7 grid. Data was sampled at approximately 50 Hz. Any text or videos displayed to the participants was shown on the monitor. \wenzhen{Those specs for the sensor should be in section III}
\dakarai{I think this information should be stated in both}

\subsubsection{Data Collection}
To create our dataset we collected gesture data from various participants. Participants watched a video of the gesture being performed on a teddy bear. A stuffed animal was used for the visualization instead of using the sensor attached to the robot arm to avoid repeating exact hand positions seen in the video. Videos were used instead of displaying text to the participant to ensure each person interpreted the gesture properly. We noticed during some preliminary trials that the participants would interpret and perform the gestures differently which is why we included the demonstration.   

In our dataset, the assignment of participants to either the training or test set determined the number of trials they performed for each gesture. Each participant’s data was entirely allocated to either the training or test set to prevent overlap. Participants in the training set performed each gesture 15 times, with nine trials conducted on the robot's upper arm and six on the lower arm. In contrast, participants in the test set performed each gesture five times, with three trials on the upper arm and two on the lower arm. We collect more data on the upper-arm because, without guidance, participants tended to favor this area. To ensure balanced data collection from the lower arm, we adjusted our procedure to explicitly tell volunteers where to touch on the arm. The order of gestures was randomized, but repeated gestures were performed consecutively. Sixteen participants (50 percent male, 50 percent female, aged 19 to 28, M = 23.12, SD = 2.18) contributed to the dataset creation.

Our dataset consisted of 1,080 samples, with 900 allocated to the training set and 180 to the test set. The training set included 150 gestures per classification, while the test set contained 30 gestures per classification.

\wenzhen{Use a standalone part to introduce your baseline}
\subsubsection{Gesture Recognition Results}
\begin{figure}
    \centering
    \includegraphics[width=\columnwidth]{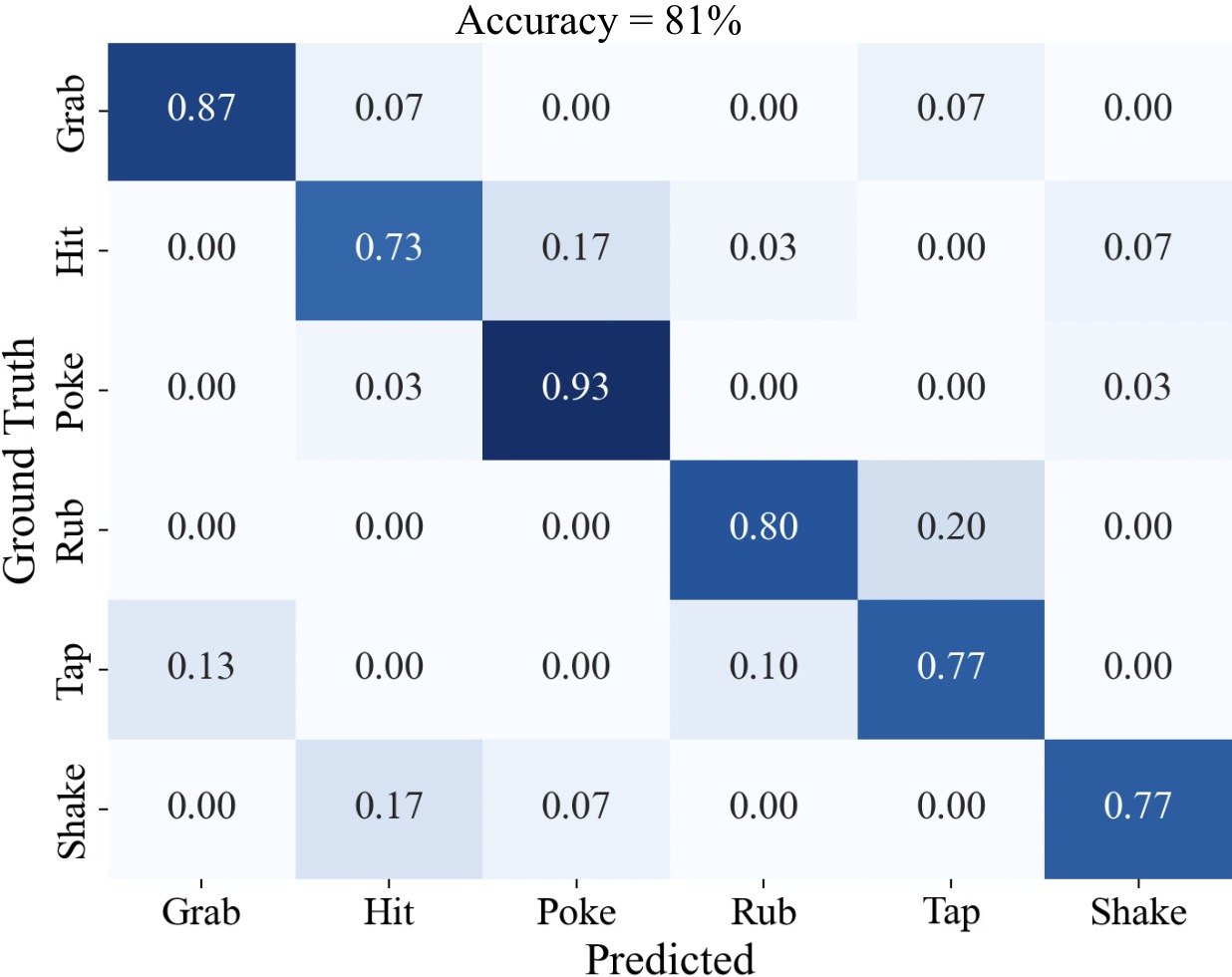}
    \caption{confusion matrix of gesture recognition result\wenzhen{Numbers in the figure are too small}}
    \label{fig:ConfusionMatrix}
\end{figure}

Our model, defined in the methods section, yielded an accuracy of 81.16\% on the test set \cref{fig:ConfusionMatrix}. Our training dataset had a size of 900. We had a 80/20 training validation split. The test set was of size 180. 

\begin{table}[h]
\caption{Models Using Number of Taxels Activated per Frame}
\label{tab:ab_study_table}
\centering
\renewcommand{\arraystretch}{1.5}
\begin{tabular}{m{10mm} p{50mm} p{10mm} }
\toprule 
\textup{\textbf{Model}}         & \textbf{Parameters}  & \textbf{Acc(\%)}\\
\midrule
\textbf{MLP (Ours)}    & Learning rate (lr): 0.00025                        & 81.16\\
LSTM            & lr: 0.0001, loss: cross entropy, optimizer: Adam   & 58.06\\
RF              & Default parameters from scikit. n\_estimators: 60  & 76.78\\
1D CNN          & lr: 0.001, loss: cross entropy, optimizer: Adam    & 71.28\\
\bottomrule
\end{tabular}
\end{table}

For a baseline we employed a variety of models (\cref{tab:ab_study_table}). Each model was trained using the training dataset split mentioned before. 

Our method was able to achieve an accuracy of 81.16\% which is higher than our highest baseline, 76.78\%, but still confused gestures for one another. Gestures "hit" and "poke" were often confused with each other, and "shake" and "grab" frequently led to similar misclassification issues as well. This is because there are times in which the signal for hit and poke are both the same. This happens when a participant would hit the sensor but they would have a bone or knuckle which is outstretched more than the rest of their fist. This means that only one or two taxels will get activated for around the same amount of time and with a similar force as poke. These results seem promising, however more data should be collected on how the system performs when people perform the same gesture differently. 

\subsubsection{Ablation Study-Feature extraction}
To ensure optimal processing of our data, we explored additional feature extraction methods. To reduce sensor noise, we applied a moving average filter with a window size of 3 to smooth the voltage readings from each taxel. 

One feature we calculated was the principal frequency for each taxel. First, for a given taxel, the minimum voltage value across all frames was subtracted from the readings. We then computed the Fast Fourier Transform (FFT) on the adjusted data and extracted the frequency with the highest magnitude. This process was repeated for each taxel, resulting in 63 features. Additionally, we calculated the standard deviation and mean for each taxel, yielding another 63 features each. To capture temporal aspects, we padded or clipped the data to ensure a consistent length of 150 frames. We identified the taxel with the highest overall magnitude and used its voltage readings across all frames as another feature.

\begin{figure}
    \centering
    \includegraphics[width=7cm]{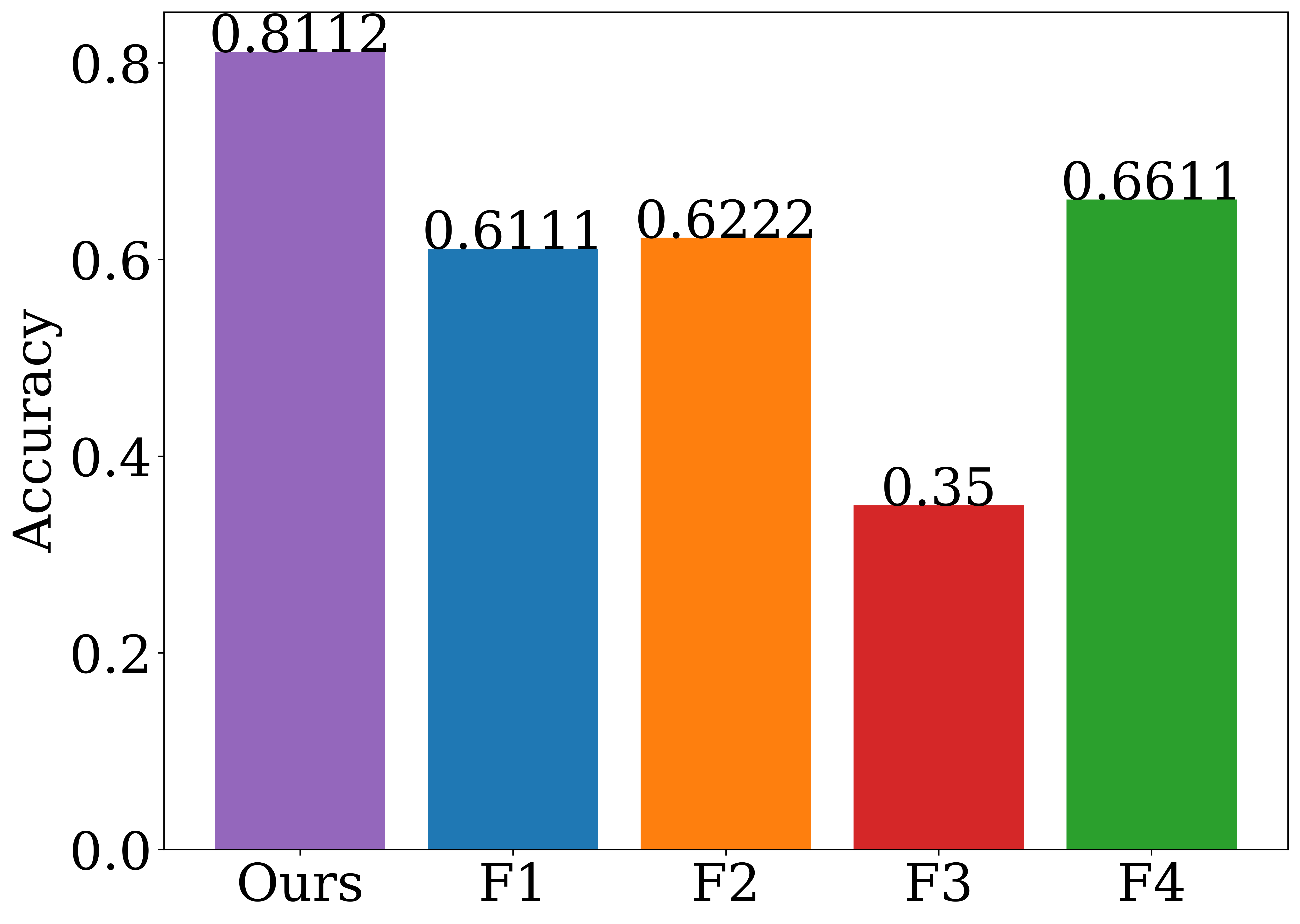}
    \caption{These are the result from training a MLP using each feature. Ours: The number of taxels activated per frame, F1: Change in senor reading for taxel with highest mean per frame, F2: The principle frequency taken from FFT, F3: Mean of each taxel over time, F4: the standard deviation of each taxel over time}
    \label{fig:MLPfeaturesResults}
\end{figure}

We trained our MLP model using four additional features and the results are shown in \cref{fig:MLPfeaturesResults}. Although the confusion matrix in \cref{fig:ConfusionMatrix} remained similar, the overall accuracy decreased. The model continued to misclassify gestures such as "hit" and "poke," and the confusion between "grab" and "shake" also increased. Performance variations were due to subtle differences in how users performed these gestures, where some features could not be adequately captured. Among the features, using the principal frequency yielded the poorest results, slightly above random guessing.

\section{CONCLUSIONS} 

In this work, we introduced a system capable of distinguishing between different types of gestures, demonstrating the potential of textiles as a medium for human-robot interaction. However, this work is still in its early stages and has several limitations. Currently, our dataset is composed of able-bodied young adults performing gestures, but people with differing abilities may interact with the system in different ways. The gestures performed by different individuals may lead to different types of signals produced by the sensor which could lead to misclassification. Additionally, our approach was primarily centered on recognizing the physical gesture itself, without fully addressing the meaning or context behind these gestures.

In future work, we plan to expand our study to include a more diverse population, incorporating individuals with various abilities. This will help us better understand how people of differing abilities interact with the system and how gesture variations can inform the design of a more robust and inclusive sensor. We aim to use this insight to refine the system's ability to distinguish between gestures across diverse user groups, ensuring more accurate recognition and responsiveness.

Misclassifying gestures could lead to the robot misunderstanding the intended message and reacting inappropriately for the given context. To address the current limitations in gesture classification, we will explore different models capable of processing both spatial and temporal data. Furthermore, we will place greater emphasis on understanding the meaning behind the gestures, allowing for more communication between users and the robot.

While this paper has served as an initial step in exploring the potential of textiles as an interactive medium, there is still much to be done. Additional future studies will focus on understanding how gestures vary across different individuals and how these variations impact interaction with robots. Once a more robust sensor system is in place, we hope to conduct open studies to further explore these interactions and refine the design to meet the needs of a broader range of users.

{\small
\bibliographystyle{IEEEtran}
\bibliography{IEEEabrv, bib}
}

\end{document}